\documentclass[journal]{IEEEtran}
\usepackage{amsmath,graphicx}
\usepackage{color,hyperref}
\usepackage[table,xcdraw]{xcolor}

\usepackage{times}
\usepackage{epsfig}
\usepackage{graphicx}
\usepackage{amsmath}
\usepackage{amssymb}

\usepackage{algorithm}
\usepackage{algorithmicx}
\usepackage{algpseudocode}
\usepackage{multirow}
\usepackage{booktabs} 
\usepackage{url}
\usepackage{cite}
\usepackage{bm}
\usepackage{booktabs}
\usepackage{makecell}
\usepackage{xcolor}

\newcommand{\R}{\mathbb{R}}

\newlength\savedwidth

\usepackage{hyperref}
\hypersetup{
    colorlinks=true,
    linkcolor=blue,
    filecolor=blue,
    urlcolor=blue
    }

\begin{document}
%
\title{From Easy to Hard: Learning Language-guided Curriculum for Visual Question Answering on Remote Sensing Data}
\author{Zhenghang Yuan,~\IEEEmembership{Student Member,~IEEE,} Lichao Mou,
	Qi Wang,~\IEEEmembership{Senior Member,~IEEE,}\\
	and Xiao Xiang Zhu,~\IEEEmembership{Fellow,~IEEE}

\IEEEcompsocitemizethanks{This work is jointly supported by the China Scholarship Council, by the European Research Council (ERC) under the European Union's Horizon 2020 research and innovation programme (grant agreement No. [ERC-2016-StG-714087], Acronym: \textit{So2Sat}), by the Helmholtz Association
through the Framework of Helmholtz AI (grant  number:  ZT-I-PF-5-01) - Local Unit ``Munich Unit @Aeronautics, Space and Transport (MASTr)'' and Helmholtz Excellent Professorship ``Data Science in Earth Observation - Big Data Fusion for Urban Research'' (grant number: W2-W3-100), by the German Federal Ministry of Education and Research (BMBF) in the framework of the international future AI lab ``AI4EO -- Artificial Intelligence for Earth Observation: Reasoning, Uncertainties, Ethics and Beyond'' (grant number: 01DD20001), and by the German Federal Ministry of Economics and Technology in the framework of the ``national center of excellence ML4Earth'' (grant number: 50EE2201C). 

Z. Yuan is with the Data Science in Earth Observation (SiPEO; former: Signal Processing in Earth Observation), Technical University of Munich (TUM), 80333 Munich, Germany. (e-mail: zhenghang.yuan@tum.de)

L. Mou and X. Zhu are with the Remote Sensing Technology Institute (IMF), German Aerospace Center (DLR), 82234 We{\ss}ling, Germany, and also with the Data Science in Earth Observation (SiPEO; former: Signal Processing in Earth Observation), Technical University of Munich (TUM), 80333 Munich, Germany. (e-mail: lichao.mou@dlr.de; xiaoxiang.zhu@dlr.de)

Q. Wang is with the School of Artificial Intelligence, Optics and Electronics (iOPEN), Northwestern Polytechnical University, 710072 Xi'an, China. (e-mail: crabwq@gmail.com)
}
}


\maketitle

\begin{abstract}
Visual question answering (VQA) for remote sensing scene has great potential in intelligent human-computer interaction system. Although VQA in computer vision has been widely researched, VQA for remote sensing data (RSVQA) is still in its infancy. There are two characteristics that need to be specially considered for the RSVQA task. 1) No object annotations are available in RSVQA datasets, which makes it difficult for models to exploit informative region representation; 2) There are questions with clearly different difficulty levels for each image in the RSVQA task. Directly training a model with questions in a random order may confuse the model and limit the performance. To address these two problems, in this paper, a multi-level visual feature learning method is proposed to jointly extract language-guided holistic and regional image features. Besides, a self-paced curriculum learning (SPCL)-based VQA model is developed to train networks with samples in an easy-to-hard way. To be more specific, a language-guided SPCL method with a soft weighting strategy is explored in this work. The proposed model is evaluated on three public datasets, and extensive experimental results show that the proposed RSVQA framework can achieve promising performance. Code will be available at {https://gitlab.lrz.de/ai4eo/reasoning/VQA-easy2hard}.

\end{abstract}

\section{Introduction}
\label{sec:intro}
\IEEEPARstart{I}MAGES from spaceborne and airborne platforms usually cover large-scale geographical areas and provide important data bases for many Earth observation (EO) applications \cite{zhu2017deep}. With the development of EO technology, there have been an increasing number of works on remote sensing image analysis, such as land use classification \cite{talukdar2020land, castelluccio2015land}, object detection \cite{cheng2016survey}, \cite{li2020object}, road extraction \cite{cheng2017automatic}, and change detection \cite{yuan2018robust}, \cite{wang2018getnet, mou2018learning}. However, due to the specialised nature of remote sensing tasks, the ability to carry out such tasks is limited to experts in the related fields. The obtained semantic information from some remote sensing tasks is not intuitive to common users, which makes it difficult to deliver the image information to users in domain-specific applications.

Fortunately, novel tasks such as image captioning \cite{xu2015show}, \cite{chen2020say} and visual question answering (VQA) \cite{shih2016look}, \cite{do2020multiple} have recently been explored for visual data. These tasks take both natural language and imagery as inputs and output easy-to-understand text in natural language. Among them, VQA has become a hot research topic in artificial intelligence community \cite{zhang2020multimodal}. Given an input image and a natural language question, VQA aims to generate a textual answer to the question based on image content \cite{antol2015vqa}. It is an interdisciplinary research area between computer vision and natural language processing \cite{lopez2017deep}. It is also a challenging task that requires a model to jointly learn multi-modal representation from both imagery and language data. More specifically, a VQA model needs to learn visual representation to understand the input image and effective features for natural language to gain an answer conditioned on image content \cite{wu2017visual}.

\begin{figure}
	\centering
	\includegraphics[width=0.5\textwidth]{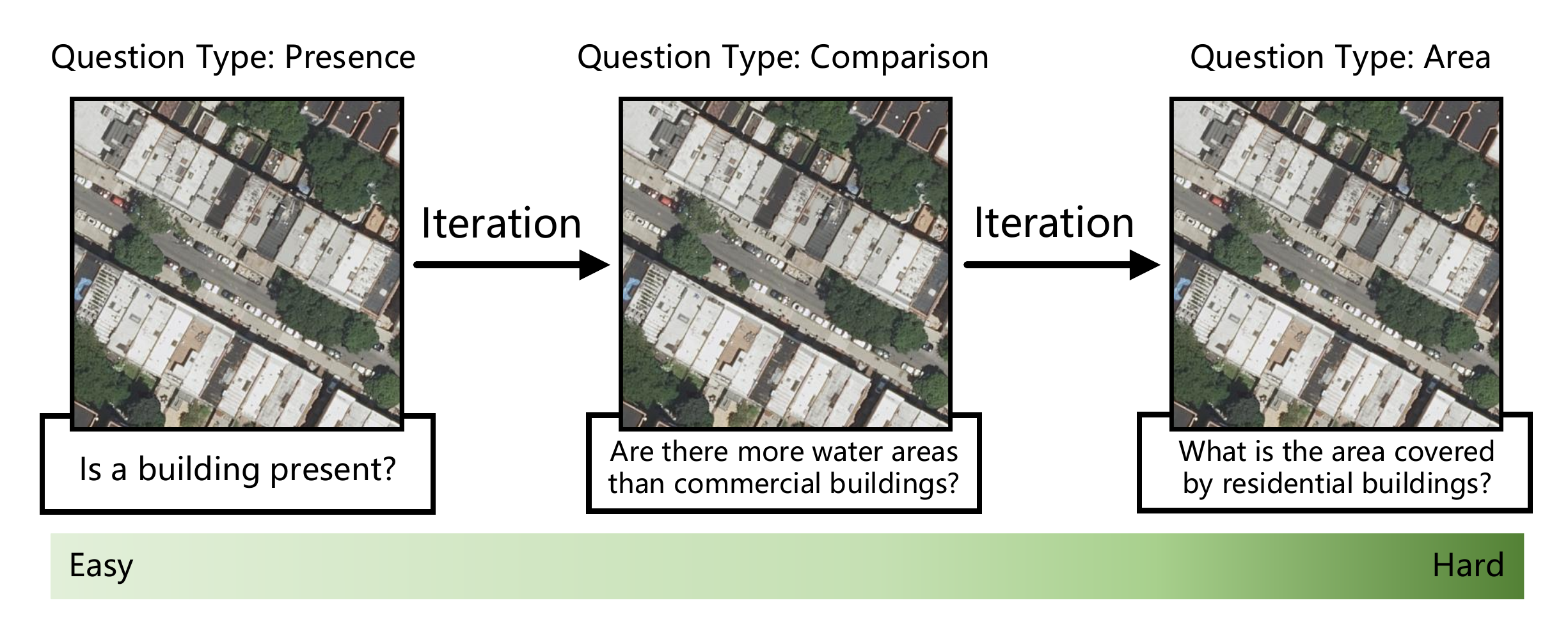}
	\caption{Motivation of the proposed method: learning features from easy samples to hard ones.}
	\label{motivation}
\end{figure}

\begin{figure*}
	\centering
	\includegraphics[width=0.95\textwidth]{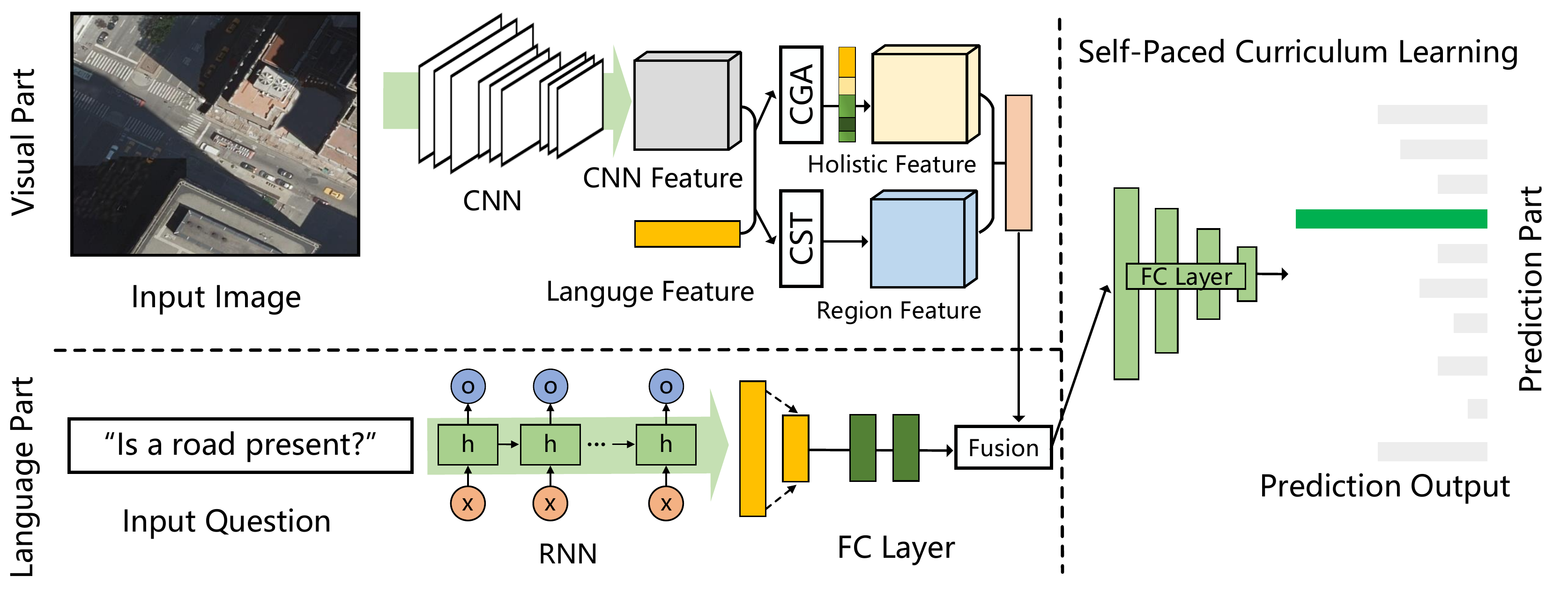}
	\caption{Main architecture of the proposed VQA method. 1) Firstly, multi-modal features are extracted from the two types of inputs, including visual features from the given image and language features from the question; 2) Then, visual features and language features are somehow fused to get the multi-modal representation; 3) Finally, the answer is predicted via a classifier.}
	\label{arch}
\end{figure*}

As for remote sensing data, VQA enables end-users to better understand a complicated remote sensing image and has great potential in human-computer interaction applications \cite{yuan2021self}. A pioneer work can be found in \cite{lobry2020rsvqa}, where the authors created two datasets and proposed a baseline model of VQA for remote sensing data (RSVQA). Although VQA in computer vision has been widely studied \cite{zhu2016visual7w, yu2017multi,zhou2020unified,peng2020answer}, VQA for remote sensing imagery is still in its infancy. Due to different image characteristics, VQA methods for natural images may not work well on remotely sensed images. Specifically, two main challenges for the RSVQA task are summarized as follows.

\begin{itemize}
	\item No object annotations available in RSVQA datasets. In computer vision, VQA models are able to make use of existing object annotations to learn features tailored to objects, which helps a lot to improve performance \cite{yang2020object,zhang2018learning}. In contrast, there are no object annotations available in RSVQA datasets, making it difficult for RSVQA models to take advantage of informative region information.
	\item Questions of different difficulty levels. Questions about each remote sensing image have significantly different difficulty levels. Fig. \ref{motivation} shows several examples. Usually, upon learning a VQA model, training image-question-answer triplets get shuffled. By doing so, the model can see different samples in no particular order and learn the task evenly without getting stuck in local optima. However, this may confuse the model and hence affect the final performance, as easy and difficult questions are in the same batch.
\end{itemize}

Aiming at the above-mentioned two challenges, our motivations are explained from two aspects. First, both holistic and region features should be well exploited to enhance visual representation for RSVQA. Though the holistic feature provides the global information of the input image, it may neglect some important details, whereas the region feature can provide more detailed semantic information, which is critical for answering complicated questions. Moreover, due to the fact that remote sensing images usually contain objects of various scales, using region representation is also helpful for addressing the scale-variation problem. To harness both two features, we propose a multi-level visual feature learning method. Specifically, the language-guided holistic image feature and the region feature are jointly learned to improve the performance of RSVQA models.

Second, the model should be trained in ascending order of learning difficulty. We humans tend to learn from easy to hard. Inspired by the human learning process, self-paced curriculum learning (SPCL) is explored in this work and shows promising results. It considers question attributes and model feedback to dynamically adjust the question sequence for model training in ascending order of difficulty \cite{spcl,sachan2016easy}, namely, from easy samples to hard ones. Albeit successful in many problems, SPCL for the RSVQA task still remains under explored.

To sum up, the main contributions of this work can be summarized as follows:

\begin{itemize}
	\item A multi-level visual feature learning method is proposed to jointly exploit both holistic and region features. Specifically, a cross-modal global attention (CGA) module is devised to learn the language-guided holistic image feature, and a cross-modal spatial transformer (CST) module is developed to learn the question-related region feature.
	
	\item The proposed CST module applies affine transformation to visual features to automatically crop informative regions without object annotations. Moreover, the language feature is also used as guidance to generate multiple spatial transformation parameters for obtaining richer region features.
	
	\item A language-guided SPCL method with a soft weighting strategy is devised for RSVQA. It takes question length and type as prior knowledge and dynamically adjusts question sequence to enable a more effective training process: learning with easy questions and then with hard ones.

\end{itemize}

The rest of the paper is organized as follows. Related works about VQA for both natural images and remote sensing images are introduced in Section \ref{Related Work}. The methodology is described in Section \ref{Methodology}. Section \ref{Experiments} presents experimental results and discussion. Finally, this paper is concluded in Section \ref{Conclusion}.

\section{Related Work}
\label{Related Work}

Multi-modal feature learning \cite{xiong2020msn,han2021multi,xiong2021ask} plays an important role in both remote sensing and computer vision tasks. For a typical VQA framework, learning multi-modal representation is also one of the core components.
Mateusz \emph{et al}. \cite{malinowski2014multi} combined semantic segmentation of scenes and symbolic reasoning over questions to learn multi-modal features. With the development of deep learning, convolutional neural networks (CNNs) and recurrent neural networks (RNNs) are usually employed for visual and language feature encoding and become mainstream feature learning methods \cite{wu2017visual, kim2019improving, huang2019multi}. Stanislaw \emph{et al}. \cite{antol2015vqa} introduced the task of free-form and open-ended VQA and employed VGGNet and LSTM to extract multi-modal features. 

Besides, visual attention mechanism is also widely used in VQA tasks to make the model focus on important pixels. Chen \emph{et al}. proposed a language-guided attention method that projects question embeddings into a visual space and to learn multi-modal features \cite{chen2015abc}. Fukui \emph{et al}. \cite{fukui2016multimodal}, Kimet al. \cite{kim2016hadamard}, and Ben \emph{et al}. \cite{ben2017mutan} designed different multimodal bilinear pooling methods to integrate visual features with language features. Yu \emph{et al}. \cite{yu2018beyond} further reduced the co-attention method into question self-attention and question-conditioned attention for learning better visual features. However, softly-attended visual features are still holistic representations of the image. The detailed region information is neglected by these methods, which is critical for alleviating the scale-variation problem in the RSVQA task.

To leverage the object-level semantic information of the input image, patch-based and object-based feature learning methods are proposed to extract more representative local features. For VQA models in computer vision, object detectors are usually used to represent the image as a collection of bounding boxes \cite{ren2015faster,teney2018tips,anderson2018bottom}. Recently, vanilla grid convolutional feature maps \cite{jiang2020defense} are also proven to be effective for visual feature learning in VQA and image captioning tasks. However, these methods all rely on bounding box annotations, which are not available in RSVQA datasets.

Compared with VQA for natural imagery, the research for RSVQA is still in its early stage. The first work for VQA on remote sensing data was introduced by Sylvain et al. \cite{lobry2020rsvqa}, where an template-based automatic method was designed to build two remote sensing-oriented datasets. The image-question-answer triplets of the two datasets were constructed via the information from OpenStreetMap and pre-defined templates. They employed a CNN to extract visual features and an RNN to learn language features. After the point-wise fusion of the multimodal features, an answer is predicted by a classification task. This work paves the way for RSVQA by providing datasets and a baseline method for further research. However, due to unique characteristics of remote sensing imagery, more specific feature learning algorithms need to be investigated and explored for this task.
 
\section{Methodology}
\label{Methodology}
As we mentioned above, this work focuses on two problems. On the one hand, how can we adaptively exploit both holistic and regional visual features for answering different types of questions. On the other hand, how can we more effectively train a model with questions of different difficulty levels. The whole architecture of the proposed RSVQA framework is shown in Fig. \ref{arch}. It consists of two parts: 1) CGA and CST modules that intend to learn multi-level visual features with multiple spatial contexts; 2) SPCL-based network training, which aims to train a model in an easy-to-hard way. Following prior works \cite{lobry2020rsvqa, zheng2021mutual}, we formulate predicting the answer conditioned on the input image and question as a classification task instead of a sentence generation task. Specifically, the final answer can be selected from the class with the highest probability. In the following sections, the two parts of our RSVQA framework will be described in detail.

\subsection{Multi-level Visual Feature Learning with Multiple Spatial Contexts}

Compared with natural images, remote sensing data usually contain much richer content due to top-down views, which enables people to ask various types of questions for the same input image. Existing RSVQA datasets in \cite{lobry2020rsvqa} contain the following five types of questions, and an example is given for each type:
\begin{itemize}
\item Rural/Urban. ``Is it a rural or an urban area?" The answer to this question can be deemed as a typical binary classification task.
\item Presence. ``Is a road present?" To answer presence-related questions, a VQA model needs to predict whether there exist specific objects.
\item Comparison. ``Are there more roads than residential buildings?" We need to compare the areas or objects involved.
\item Area. ``What is the area covered by residential areas?" The model needs to seek out the target according to the question.
\item Count. ``What is the amount of small buildings?" To answer count-related questions, the numbers of specific objects need to be predicted.
\end{itemize} 

\begin{figure}[!]
	\centering
	\includegraphics[width=0.48\textwidth]{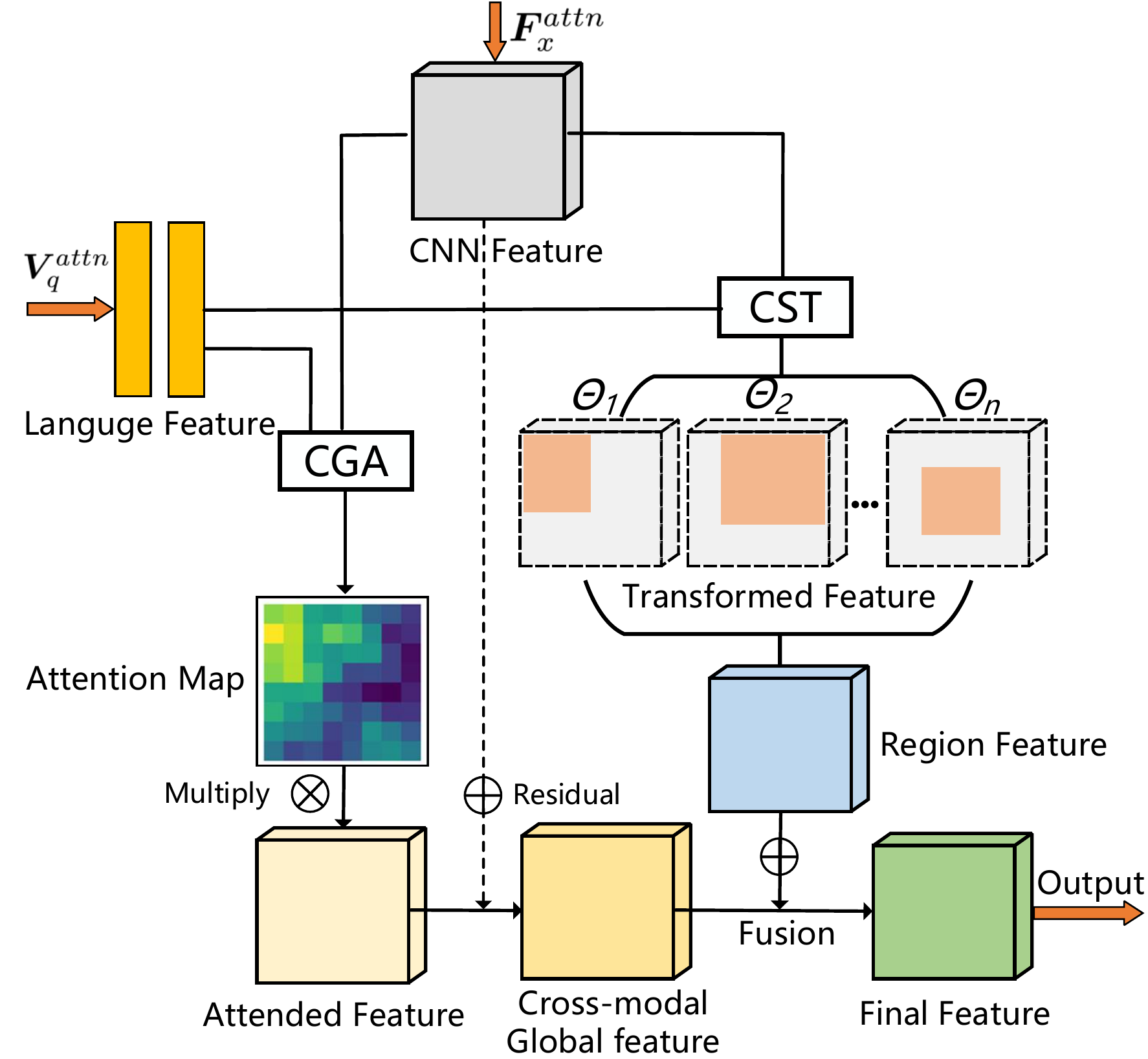}
	\caption{Illustration of the proposed CGA and CST visual feature learning modules. Using the proposed two modules, cross-modal global and spatially-transformed visual features can be learned jointly.}
	\label{visf}
\end{figure}

As can be seen, different image regions and spatial contexts need to be employed to answer various types of questions. Thus, we propose to combine holistic (global) and region (local) features to learn multi-level visual representation with multiple spatial contexts. Specifically, CGA is designed to extract global visual features with the guidance of language. CGA utilizes language features as the guidance to generate global attention maps on the whole image. Meanwhile, the CST module is proposed to extract region features in an adaptive way. CST learns to spatially transform feature maps to be of different scales and poses, and the transformed features can be used to enhance visual representations. Details of the two proposed modules are illustrated in Fig. \ref{visf} and are introduced in the following subsections.

\subsubsection{Cross-modal Global Attention Module}
Attention mechanism has shown its effectiveness in many computer vision tasks \cite{vaswani2017attention, xiong2020msn, wang2019learning}. This is mainly because focusing on some important regions of an image can improve the discriminability of visual features. For object recognition or semantic segmentation tasks, only one modality, i.e., vision, is input to the model, and self-attention is used. However, there are two modalities in our case.  Therefore, we propose a cross-modal global attention, i.e., CGA, which exploits the language feature as guidance to generate global attention maps on all locations.

Formally, let $\bm{x}$ be the input image, and the corresponding question is denoted as $\bm{q}$. For visual and language modalities, CNNs and RNNs are commonly used as feature encoders. Correspondingly, visual feature $\bm{F}_x \in \R^{N\times C\times H\times W}$ and language feature vector $\bm{v}_q \in \R^{N\times L}$ can be obtained with the networks.
$N$ is the batch size, $C$ is the number of channels, and $H, W$ represent the height and width of the input image. $L$ denotes the dimension of the language feature vector. The typical, single-modality attention mechanism first encodes the visual feature $\bm{F}_x$ into three independent features: the query $\bm{Q}$, key $\bm{K}$ and value $\bm{V}$. The key idea of attention mechanism is to assign weights to the input value $\bm{V}$ according to a similarity function. Usually, the attention weights are computed by the compatibility between the input query and the corresponding key. Generally, the attention \cite{vaswani2017attention} for the single modality input can be calculated by 
\begin{equation}
	\text{Attention}(\bm{Q,K,V}) = \text{softmax}(\frac{\bm{QK}^T} {\sqrt{d_k}})\bm{V},
\end{equation}
where $d_k$ is the number of channels of $\bm{F}_x$.

However, for RSVQA, we make use of the information of the natural language question to assign weights for visual features. To this end, we devise a multi-modal compatibility function to compute the similarities between language and visual features at different locations. Firstly, a $1\times 1$ convolution layer and a fully connected (FC) layer are employed to transform the visual and language features into $\bm{F}^{attn}_x \in \R^{N\times C\times H\times W}$ and $\bm{V}^{attn}_q \in \R^{N\times C}$. Then, we expand the dimension of $\bm{V}^{attn}_q$ to the same dimension as $\bm{F}^{attn}_x$. Afterwards, the query $\bm{Q}^{attn}$ can be computed as
\begin{equation}
	\bm{Q}^{attn}=\text{Conv}_{1\times1}(\text{ReLU}(\frac{\bm{F}^{attn}_x}{\left \| \bm{F}^{attn}_x \right \|_2}+\frac{\bm{V}^{attn}_q}{\left \| \bm{V}^{attn}_q \right \|_2})).
\end{equation}

Finally, the cross-modal attention can be defined as
\begin{equation}
	\text{Attention}(\bm{Q}^{attn},\bm{F}^{attn}_x,\bm{V}) = \text{softmax}(\frac{\bm{Q}^{attn}{\bm{F}^{attn}_x}^T} {\sqrt{d^{\prime}_k}})\bm{V}+\bm{F}_x,
\end{equation}
where the value $\bm{V}$ is computed by $\bm{V} = \text{Conv}_{1\times1}(\bm{F}_x)$, and $d^{\prime}_k$ is the channel dimension of $\bm{F}^{attn}_x$.

\subsubsection{Cross-modal Spatial Transformer Module}
Attention mechanism can make model focus on informative image features by using soft weights. However, it is still a global feature learning method with a fixed spatial context. Compared with global features, multi-level features are more effective. Therefore, we propose a cross-modal spatial transformer, i.e., our CST module, to extract features with adaptive scales and spatial contexts. The spatial transformation in CST includes cropping, translation, and scaling, which can be learned in an end-to-end manner without object annotations. As opposed to the \textit{soft attention} in CGA module, the CST module can be viewed as a \textit{hard attention} method for region feature learning. 

The spatial transformer in CST is a differentiable transformation module, and it is conditioned on both visual and language features. Specifically, there are three sub-components in the spatial transformer following the work in \cite{DBLPJaderbergSZK15}: localization network, parameterized grid sampling, and differentiable bilinear sampling.

In this work, the transformation parameter $\bm{T}_{\theta}$ can be defined as
\begin{equation}
	\bm{T}_{\theta}=
	\begin{bmatrix}
		& s_1 & 0 & t_x\\ 
		& 0 & s_2  &t_y
	\end{bmatrix},
\end{equation}
where $s_1, s_2$, and $t_x, t_y$ are parameters controlling the scaling, translation, and cropping transformation, respectively.

To predict the transformation parameter $\bm{T}_{\theta}$, we design a cross-modal localization network. In CGA, $\bm{F}^{attn}_x$ and $\bm{Q}^{attn}$ are computed for generating the cross-modal attention. In cross-modal localization network, we reuse $\bm{F}^{attn}_x$ and $\bm{V}^{attn}_q$ for predicting the transformation parameter $\bm{T}_{\theta}$. Since different feature channels focus on different parts of the image \cite{zheng2017learning,xiong2021ask}, we propose to use multiple spatial transformers from different channel groups to extract richer visual features. This can be defined as



\begin{equation}
\bm{M}^{attn}_1, \bm{M}^{attn}_2 = \text{split}(\frac{\bm{F}^{attn}_x}{\left \| \bm{F}^{attn}_x \right \|_2}+\frac{\bm{V}^{attn}_q}{\left \| \bm{V}^{attn}_q \right \|_2}).
\end{equation}
We split the cross-modal feature into $\bm{M}^{attn}_1$ and $\bm{M}^{attn}_2$ along the channel dimension evenly. Then, two transformation parameters are predicted by split features as
\begin{equation}
	\begin{split}
	    \bm{T}_{\theta_1}=\text{FC}(\text{ReLU}(\bm{M}^{attn}_1)),\\
		\bm{T}_{\theta_2}=\text{FC}(\text{ReLU}(\bm{M}^{attn}_2)),\\
	\end{split}
\end{equation}
where $\text{FC}$ denotes FC layers. $\bm{T}_{\theta_1}$ and $\bm{T}_{\theta_2}$ denote the predicted transformation parameters. 

Note that two spatial transformers share a similar differentiable bilinear sampling process, which can be defined by
\begin{equation}
\small
	\begin{split}
		\bm{E}_{i}^c = \sum\limits_u^H \sum\limits_v^W \bm{F}_{x}^c(u,v)  \text{max}(0,1 - |{x_{i}} - v|) 
		\text{max}(0,1 - |{y_{i}} - u|), 
	\end{split}
\end{equation}
where ${i \in \{1,2,...,WH\}}$ is the coordinate index, and ${c}$ is the channel index. Transformed spatial coordinates ${(x_{i},y_{i})}$, and the feature coordinates ${(u,v)}$ of $\bm{F_x}$ are normalized in the range of $[-1,1]$. $\bm{E}_{i}^c$ denotes the sampled features using the transformation parameter $\bm{T}_{\theta_1}$ or $\bm{T}_{\theta_2}$. For each spatial transformer, we compute partial derivatives for both features and coordinates. By this means, the whole networks can be trained in an end-to-end manner.


\subsection{Cross-modal Feature Learning: From Easy to Hard}

Traditional stochastic training strategy usually takes the input samples in a random order, while this is opposite to the learning process of human. Curriculum learning (CL) \cite{2009Curriculum}, self-paced learning (SPL) \cite{kumar2010self}, and SPCL \cite{spcl} are proposed as more reasonable training algorithms for machine learning models. The core idea of them is to train a model starting from easy samples and gradually including hard ones. Previous works \cite{sachan2016easy,li2020competence} show that designing proper ranking functions to organize training samples in ascending order of learning difficulty is helpful for improving the model performance. Since there exist questions with different difficulty levels for the same remote sensing image, training an RSVQA model from easy to hard is a more reasonable strategy. 

In this subsection, we propose a language-guided SPCL training method for RSVQA. Generally, SPCL is composed of two parts: SPL and CL.
SPL can be reformulated as an optimization problem, and the curriculum is dynamically adjusted according to model feedback during the training phase. The curriculum in CL is determined by prior knowledge, which is a prejudgment about the difficulty level of a specific task.

In this work, the target of SPL is to adjust the sequence of input samples during the training stage. Specifically, SPL utilizes adaptive weights for each training sample to control the training sequence by an importance sampling strategy. Let ${\bm{v}} = [{v_1},{v_2},...,{v_N}]$ denote the weight vector for each sample in $N$ training questions. $\bm{x}_{i}$ is the $i$-th input image, and $\bm{q}_{i}$ is the $i$-th input question. $g(h(\bm{x}_i),s(\bm{x}_i),\bm{q}_i; \bm{w})$ represents the whole RSVQA model. Here,  $h(\bm{x}_i)$ denotes the global feature learned by CGA module, and $s(\bm{x}_i)$ denotes the transformed feature learned by CST module. $\bm{w}$ represents the learnable network weights in the whole model. Then, SPL is exploited to train the model with samples organized in ascending order of learning difficulty. Based on the learned multi-level features, the SPL loss can be defined as 
\begin{equation}
	\begin{split}
	\min _{\bm{w}, \bm{v}} \mathbb{E}(\bm{w}, \bm{v}, \lambda)=\sum_{i=1}^{N} v_{i} L\left(y_{i}, g\left(h(\bm{x}_i),s(\bm{x}_i), \bm{q}_{i}, \bm{w}\right)\right)
	\\+f(\bm{v} ; \lambda),
	\end{split}
\label{eq}
\end{equation}
where samples with larger $v$ have larger influence on model training and vice versa. $y_i$ is the ground truth label. Since we take answer prediction as a classification task, its loss function is a cross-entropy function represented by $L\left(y_{i}, g\right)$. $\lambda$ can be interpreted as the ``age'' of the model, which is used to control the learning pace.

Actually, $f(\bm{v} ; \lambda)$ is a self-paced regularizer for controlling the learning process. The vector ${\bm{v}}$ is learnable and updated by optimizing the SPL loss function. Basically, elements of $\bm{v}$ can be hard (0 or 1) or soft (from 0 to 1). In what follows, we study the soft regularizer for SPL to enable a more flexible training of the RSVQA model. Specifically, the soft regularizer can be defined as follows:
\begin{equation}
	\begin{split}
		f=\lambda\left(\frac{1}{2} \bm{v}^{2}-\bm{v}\right), \bm{v} \in (0,1)^{N}.
	\end{split}
\end{equation}

Given that there are two disjoint blocks of variables, i.e., network weights $\bm{w}$ and weight vector $\bm{v}$, SPL is a biconvex optimization problem. Usually, the alternative convex search algorithm is used to solve it. When network parameters $\bm{w}$ including all learnable weights in CGA and CST modules are fixed, the global optimum $\bm{v^*}$ for the regularizer, $\bm{v^*}$ can be computed by
\begin{equation}
	v_{i}^{*}=\left\{\begin{array}{ll}
		-\frac{L}{\lambda}+1, & \text { if } L\left(y_{i}, g\left(h(\bm{x}_i),s(\bm{x}_i), \bm{q}_{i}, \bm{w}\right)\right) \le \lambda, \\
		0, & \text { otherwise. }
	\end{array}\right.
\end{equation}

After vector $\bm{v}$ is updated, we fix $\bm{v}$ and optimize network weights $\bm{w}$ by a stochastic gradient descent (SGD) optimizer. Since easy samples can be quickly fitted with limited iterations, the loss values for easy samples are usually smaller than those for hard ones. So if the loss value $L$ is not larger than $\lambda$, the corresponding input question will be taken as an easy sample and trained with high priority. Otherwise, $v_{i}^{*}$ is set to 0, and the corresponding question will not be used for training. 

As we mentioned, $\lambda$ is the ``age'' of the model that increases gradually along with the training iteration. In this work, we record the maximum and minimum loss values of epoch $t-1$, and use them to update $\lambda$ as follows:
\begin{equation}
	\lambda=(\max (L^{t-1})-\min (L^{t-1}))\cdot K+\min (L^{t-1}),
\end{equation}
where $K$ is used to adjust the value of $\lambda$. Specially, we define $K$ as a dynamic changing parameter for controlling the learning pace. The initial value of $K$ is set to 0.5, and it is updated during the training stage by
\begin{equation}
    K=0.5+\frac{{t}}{15} \times 0.1.
\end{equation}
When the value of $\lambda$ increases, the model includes more difficult questions with larger loss values.

However, SPL does not incorporate prior knowledge in the learning process. In the initial stage, the network weights are still randomly initialized, and the loss values of easy and hard examples may not be accurate to determine the true difficulty order. Thus, incorporating prior knowledge is necessary in our case. Inspired by CL, we design a curriculum, namely, a ranking function to organize questions in an easy-to-hard order at the beginning of training. By combining SPL with CL, the proposed language-guided SPCL method for RSVQA can take advantages of these two learning regimes.

Two factors are considered in this work to design the ranking function in CL: question length and question type. In most cases, longer questions are usually more complicated than shorter ones. In addition, different types of questions also have different difficulty levels. For example, object recognition is usually easier than counting task.

Based on this prior knowledge, the SPCL loss can be defined as:
\begin{equation}
	\begin{split}
		\min _{\bm{w}, \bm{v}} \mathbb{E}(\bm{w}, \bm{v}, \lambda, \Psi)=\sum_{i=1}^{N} v_{i} L\left(y_{i}, g\left(h(\bm{x}_i),s(\bm{x}_i), \bm{q}_{i}, \bm{w}\right)\right) \\
		+f(\bm{v} ; \lambda), 
		\text { s.t. } \bm{v} \in \Psi,
	\end{split}
	\label{spc}
\end{equation}
where $\Psi=\left\{\bm{v} \mid \bm{a}^{T} \bm{v} \leq c\right\}$ is a pre-defined curriculum region to initialize the weight vector $\bm{v}$. $c$ is a constant, and $\bm{a}$ is the ranking function that indicates the difficulty levels of training samples.
Usually, $\Psi$ can be derived from a task-specific ranking function $\bm{a}$ and a constant $c$. In this work, $\bm{a}$ is defined by calculating $a_i=W^q_iQ^q_i$ for the $i$-th question. $W^q_i$ denotes the pre-defined prior weight for different question types. $Q^q_i$ is the length of question, which is normalized by dividing the maximum question length. Note that CL is only used in the initial stage, and SPL adaptively updates $\bm{v}$ in the rest of the training stage.

\section{Experiments}
\label{Experiments}

\subsection{Datasets}
In order to evaluate the proposed RSVQA framework, we conduct experiments on three public datasets. Two of these are released by \cite{lobry2020rsvqa}: the low resolution (LR) and high resolution (HR) RSVQA datasets. LR dataset is based on Sentinel-2 images at 10 m resolution. It contains 772 images of size 256×256 and 77,232 question-answer pairs. Among them, there are 23,002 (29.78\%) pairs for the count question type, 22,882 (29.63\%) for presence, 30,576 (39.59\%) for comparison, and 772 (1.00\%) for rural/urban. Overall, 77.8\%, 11.1\%, and 11.1\% of original tiles are divided into the training set, validation set and test set, respectively. HR dataset is collected from high resolution orthoimagery data at 15 cm resolution. It consists of 10,659 images of size 512×512 and a total number of 1,066,316 question-answer pairs. Specifically, there are 277,702 (26.04\%) pairs for count, 278,335 (26.10\%) for presence, 353,772 (33.18\%) for comparison, and 156,507 (14.68\%) for area. In general, 61.5\%, 11.2\%, 20.5\%, and 6.8\% of original tiles are split into the training set, validation set, test set 1, and test set 2, respectively.

Another dataset is the RSIVQA dataset proposed in \cite{zheng2021mutual}. It is created on top of five existing datasets including UC-Merced (UCM) \cite{yang2010bag}, Sydney \cite{zhang2014saliency}, AID \cite{xia2017aid}, HRRSD \cite{zhang2019hierarchical}, and DOTA \cite{xia2018dota}. RSIVQA utilizes two types of annotation methods: manual annotation and automatic generating. In total, this dataset contains more than 110,000 VQA triplets. According to the answer type, these triplets can be divided into three types: yes/no, number, and others. Following the experimental setting described in \cite{zheng2021mutual}, we randomly sample 80\%, 10\%, 10\% of all triplets as the training set, validation set and test set, respectively.

\subsection{Implementation Details}
We use Adam optimizer with an initial learning rate of 1e-5 for model training. The batch size is set to 280 for LR and 70 for HR dataset. For methods without multi-level feature learning, we make use of 150 epochs and 35 epochs to train the models on the LR and HR datasets, respectively. Since more epochs are needed for models with the multi-level feature learning module to converge, 300 epochs on the LR dataset and 70 epochs on the HR dataset are used for these models.

We take the method in \cite{lobry2020rsvqa} as the baseline model and conduct experiments to demonstrate the effectiveness of the proposed method. Note that the same language feature embedding network is used in both the baseline and the proposed framework. As for the visual feature learning module, we employ the proposed CST and CGA modules. Additionally, the traditional cross entropy loss is replaced by the proposed SPCL loss function to enable the easy-to-hard learning strategy. To comprehensively evaluate the proposed modules, we exploit accuracy with respect to question type, average accuracy, and overall accuracy as evaluation metrics. Each model is trained 3 times in every experiment (except for the results in Table \ref{tabel-6}), and the mean and standard deviation are reported on both datasets.

\begin{table}
	\centering
	 \caption{Comparisons on the LR Dataset. Both the Mean Value and the Standard Deviation are Reported.}
	\scalebox{0.92}{
		\begin{tabular}{m{2.7cm}<{\centering} m{2.7cm}<{\centering} m{2.8cm}<{\centering}}
			\toprule
			Types            & Baseline      & SPCL+MLL           \\  \midrule
			Count            & 67.01\% (0.59\%)    & \textbf{69.22\%} (0.33\%)   \\
			Presence         & 87.46\% (0.06\%)    & \textbf{90.66\%} (0.24\%) \\ 
			Comparison       & 81.50\% (0.03\%)    & \textbf{87.49\%} (0.10\%) \\ 
			Rural/Urban      & 90.00\% (1.41\%)    & \textbf{91.67\%} (1.53\%) \\ 
			Average Accuracy & 81.49\% (0.49\%)    & \textbf{84.76\%} (0.35\%) \\ 
			{Overall Accuracy} & 79.08\% (0.20\%)    & \textbf{83.09\%} (0.15\%) \\  \bottomrule
		\end{tabular}
	}
	\label{tabel-1}
\end{table}

\begin{table}
	\centering
	\caption{Comparisons on the Test set 1 of the HR Dataset. Both the Mean Value and the Standard Deviation are Reported.}
	\label{tabel-2}
	\scalebox{.90}{
		\begin{tabular}{m{2.1cm}<{\centering} m{1.9cm}<{\centering} m{1.9cm}<{\centering} m{2.1cm}<{\centering}}
			\toprule
			Types            & Baseline         & SPCL & SPCL+MLL \\ \midrule
			Count            & 68.63\% (0.11\%) & {68.91\%} (0.03\%)  & \textbf{69.06\%} (0.13\%)    \\
			Presence         & 90.43\% (0.04\%) & {90.66\%} (0.08\%) & \textbf{91.39\%} (0.15\%)    \\
			Comparison       & 88.19\% (0.08\%) & {89.07\%} (0.27\%)  & \textbf{89.75\%} (0.12\%)  \\
			Area             & 85.24\% (0.05\%) & {85.66\%} (0.26\%) & \textbf{85.92\%} (0.19\%)   \\
			Average Accuracy & 83.12\% (0.03\%) & {83.57\%} (0.11\%)  & \textbf{83.97\%} (0.06\%)  \\
			{Overall Accuracy} & 83.23\% (0.02\%) & {83.69\%} (0.11\%)  & \textbf{84.16\%} (0.05\%)  \\ 
			\bottomrule
		\end{tabular}
	}
\end{table}

\begin{table} [!]
	\centering
	\caption{Comparisons on the Test set 2 of the HR Dataset. Both the Mean Value and the Standard Deviation are Reported.}
	\label{tabel-3}
	\scalebox{.86}{
		\begin{tabular}{m{2.1cm}<{\centering} m{2cm}<{\centering} m{2cm}<{\centering} m{2.4cm}<{\centering}}
			\toprule
			Types            & Baseline         & SPCL & SPCL+MLL \\ \midrule
			Count            & 61.47\% (0.08\%) & {61.63\%} (0.11\%)  & \textbf{61.95\%} (0.08\%)    \\
			Presence         & 86.26\% (0.47\%) & {87.10\%} (0.12\%) & \textbf{87.97\%} (0.06\%)    \\
			Comparison       & 85.94\% (0.12\%) & {86.93\%} (0.23\%)  & \textbf{87.68\%} (0.23\%)  \\
			Area             & 76.33\% (0.50\%) & {76.81\%} (0.14\%) & \textbf{78.62\%} (0.23\%)   \\
			Average Accuracy & 77.50\% (0.29\%) & {78.12\%} (0.13\%)  & \textbf{79.06\%} (0.15\%)  \\
			Overall Accuracy & 78.23\% (0.25\%) & {78.25\%} (0.11\%)  & \textbf{79.29\%} (0.15\%)  \\ 
			\bottomrule
		\end{tabular}
	}
\end{table}

\subsection{Comparisons on the LR Dataset}
The proposed framework consists of two main components, namely multi-level visual feature learning and SPCL. The numerical results on the LR dataset are shown in Table \ref{tabel-1}. In this table, SPCL denotes SPCL with the soft regularizer, which is used as our final loss. SPCL+MLL represents the full-fledged framework proposed in this paper. MLL stands for multi-level visual feature learning and includes both CGA and CST modules.

The LR dataset contains four types of questions: rural/urban, presence, comparison and count. Taking into account the difficulty level of each question type, we set the prior weights $W^q_i$ as $\{\text{rural/urban: 1.0}, \text{presence: 1.0},$ $\text{comparison: 3.0}, \text{count: 4.0}\}$. From Table \ref{tabel-1}, we can see that the proposed method gains better results. An improvement of 3.27\% in average accuracy can be achieved. In addition, the proposed method improves overall accuracy by 4.01\%. We also find that much greater performance is obtained for all question types. Especially for comparison type, there is an improvement of about 6\% in accuracy compared with the baseline method. These results indicate that multi-modal visual feature learning and SPCL training strategies can enhance the performance of RSVQA.

\begin{table}
	\centering
	\caption{Comparisons on the RSIVQA Dataset. Both the Mean Value and the Standard Deviation are Reported.}
	\scalebox{0.92}{
		\begin{tabular}{m{2.7cm}<{\centering} m{2.7cm}<{\centering} m{2.8cm}<{\centering}}
			\toprule
			Types            & MAIN \cite{zheng2021mutual}      & SPCL+MLL           \\  \midrule
			Yes/No            & 92.82\% (0.56\%)    & \textbf{95.49\%} (0.29\%)   \\
			Number         & \textbf{56.71\%} (2.92\%)    & 49.03\% (0.36\%) \\ 
			Others       & 54.50\% (3.06\%)    & \textbf{63.65\%} (2.92\%) \\ 
			Overall Accuracy & 77.39\% (1.27\%)    & \textbf{79.70\%} (0.43\%) \\ 
			\bottomrule
		\end{tabular}
	}
	\label{tabel-4}
\end{table}

\begin{table*}
	\centering
	\caption{Ablation Study on the LR Dataset. Both the Mean Value and the Standard Deviation are Reported.}
	\label{tabel-5}
	\scalebox{1.16}{
		\begin{tabular}{m{2.4cm}<{\centering} m{2.4cm}<{\centering} m{2.5cm}<{\centering} m{2.5cm}<{\centering} m{2.4cm}<{\centering}}
			\toprule
			Types            & Baseline      & CGA Module & CST Module  & MLL          \\ \midrule
			Count            & 67.01\% (0.59\%)    & {67.84\%} (0.23\%)     & 68.44\% (0.17\%)   & \textbf{68.53\%} (0.10\%)   \\
			Presence         & 87.46\% (0.06\%)    & 89.49\% (0.58\%)     & {89.77\%} (0.22\%) & \textbf{90.13\%} (0.49\%) \\
			Comparison       & 81.50\% (0.03\%)    & 86.62\% (0.33\%)    & \textbf{87.21\%} (0.61\%) & 86.91\% (0.50\%) \\
			Rural/Urban      & 90.00\% (1.41\%)    & 88.67\% (1.15\%)    & 89.00\% (1.00\%) & \textbf{92.00\%} (1.00\%) \\
			Average Accuracy & 81.49\% (0.49\%)    & 83.15\% (0.38\%)    & {83.60\%} (0.12\%) & \textbf{84.39\%} (0.26\%) \\
			Overall Accuracy & 79.08\% (0.20\%)    & 81.96\% (0.24\%)    & 82.45\% (0.32\%) & \textbf{82.50\%} (0.18\%) \\ 
			\bottomrule
		\end{tabular}
	}
\end{table*}

\begin{figure}[t]
	\centering
	\includegraphics[width=0.47\textwidth]{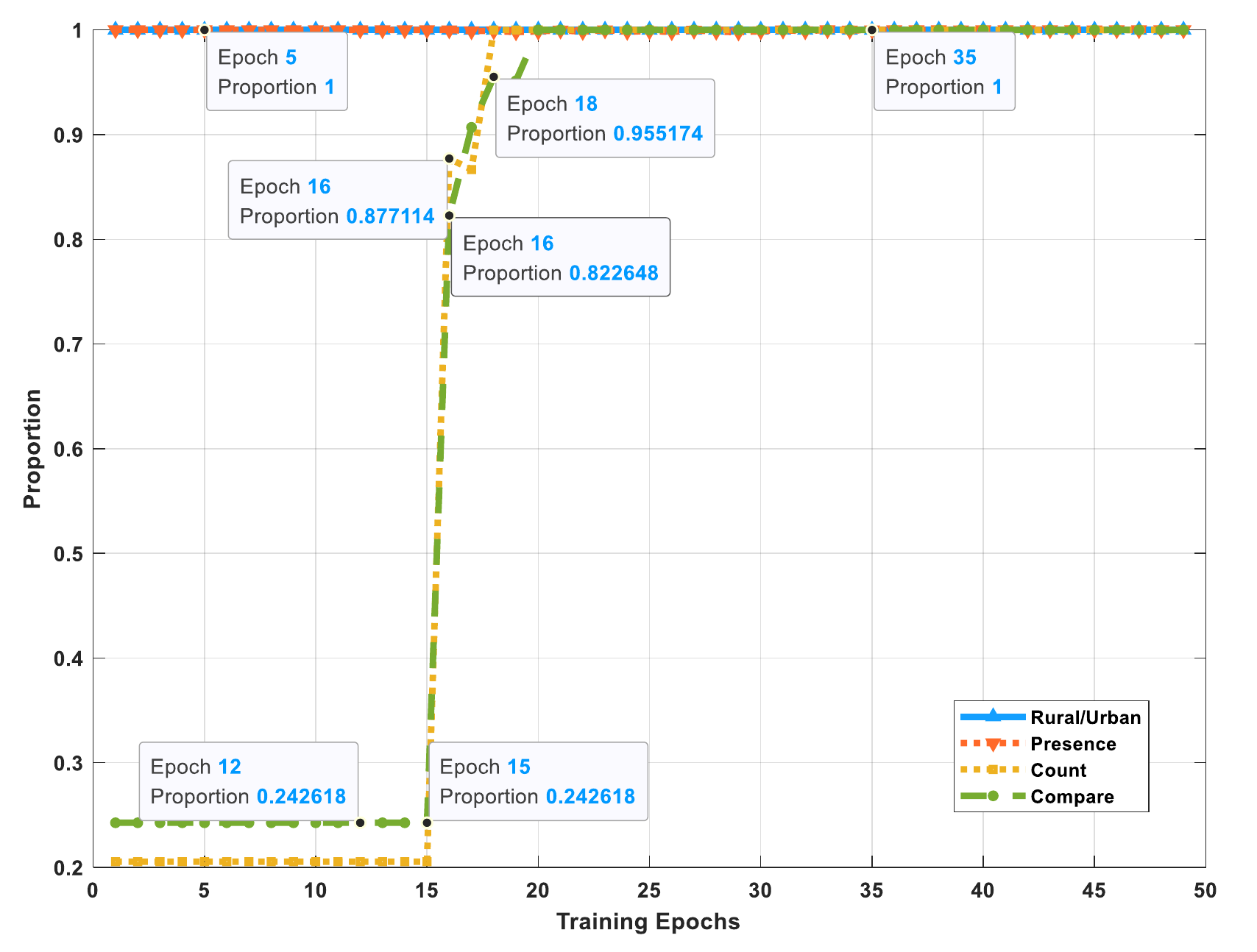}
	\caption{Visualization of the proportions of training samples for SPCL on the LR dataset. The proportions of different question types during the first 50 training epochs are displayed by different colors of lines.}
	\label{train_sample}
\end{figure}

\begin{figure}[t]
	\centering
	\includegraphics[width=0.467\textwidth]{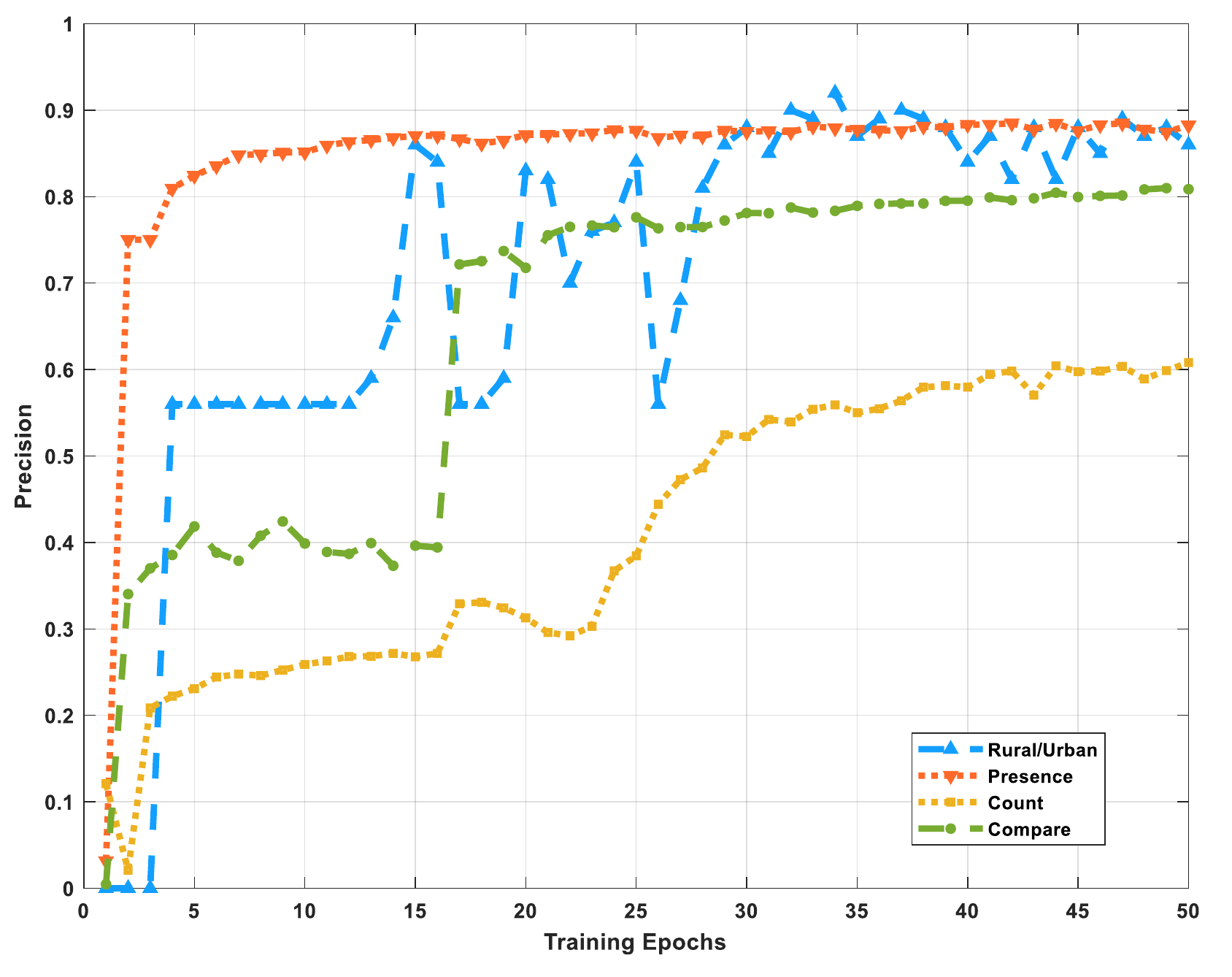}
	\caption{Visualization of the precisions of different question types for SPCL on the LR dataset. The precisions of different question types during the first 50 training epochs are displayed by different colors of lines.}
	\label{train_precision}
\end{figure}

As shown in Fig. \ref{train_sample}, the proportions of training samples for SPCL on the LR dataset are visualized. Note that this model does not include the MLL feature learning module. In this figure, the proportions of training samples for four question types are compared in detail. At the first 15 epochs, CL is used to initialize the weight vector $\bm{v}$. Thus, the proportions of count and comparison question types are obviously smaller than those of rural/urban and presence. After the first 15 epochs, SPL is used to update the vector $\bm{v}$ to control the training order of different question types. The proportion of count type is the smallest at the beginning few epochs. Then, hard examples are gradually included. The results also support our assumption on the difficulty levels of different question types. 

The precisions of different question types for SPCL during the training stage on the LR dataset are visualized in Fig. \ref{train_precision}. From this figure, it can be clearly observed that easy questions can achieve higher precisions at the first 15 epochs. Afterwards, the precisions of more difficult question types are improved rapidly. Moreover, we can see that the precisions of easy question types are not affected by adding more difficult training samples.

\begin{figure*}
	\centering
	\includegraphics[width=0.95\textwidth]{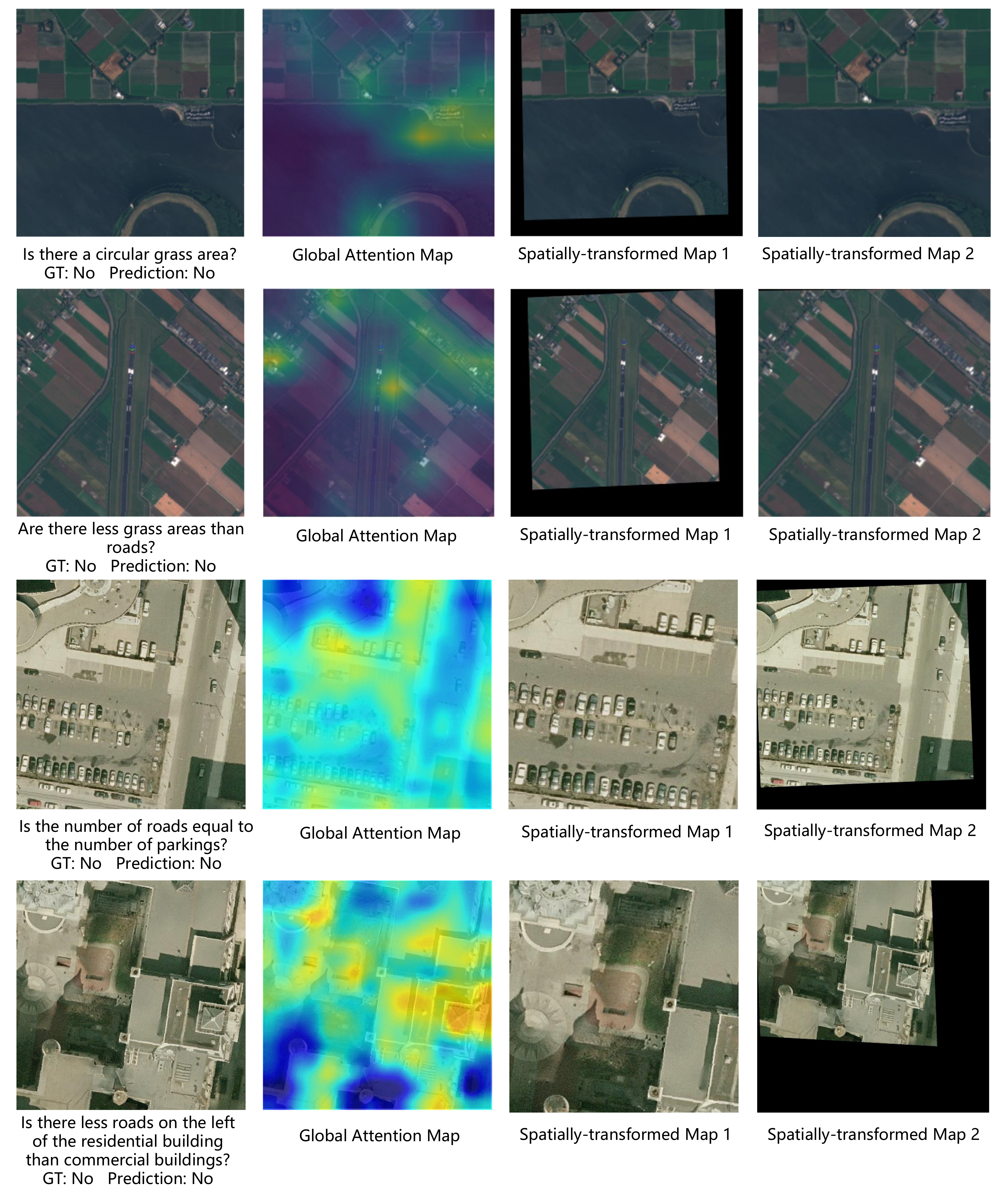}
	\caption{Illustration of some qualitative examples on the LR (the upper two rows) and HR (lower two rows) datasets. The first column is the input image. The corresponding questions and predicted answers are presented bellow each image. Global attention maps are displayed in the second column. For the last two columns, the spatial transformation is shown. (Best viewed in color.) }
	\label{visp}
\end{figure*}

The global attention maps and spatially-transformed maps are visualized in Fig. \ref{visp}. The second column of the figure shows that global attention mechanism learns to focus on important pixels of remote sensing images. The third and fourth columns indicate that spatial transformers extract visual features of local regions. Note that we apply learned transformation parameters to the original remote sensing images instead of feature maps for a clear visualization.

\begin{table*}
	\centering
	\caption{Performance comparisons of different numbers of training samples on the LR Dataset.}
	\scalebox{1.26}{
		\begin{tabular}{m{2.1cm}<{\centering} m{2.1cm}<{\centering} m{2cm}<{\centering} m{2cm}<{\centering} m{2cm}<{\centering} m{2cm}<{\centering}}
			\toprule
			Types            & 10\% Training         & 40\% Training  & 70\% Training & 100\% Training              \\ \midrule
			Count            & 61.22\%     & 63.49\%   & {67.92\%}   & \textbf{68.61\%} \\
			Presence         & 80.41\%     & 85.71\%   & 88.85\%    & \textbf{90.96\%}  \\
			Comparison       & 84.90\%     & 85.67\%   & 86.16\%    & \textbf{88.06\%}  \\
			Rural/Urban      & 76.67\%     & 82.50\%   & 90.00\%  & \textbf{92.00\%}    \\
			Average Accuracy & 75.80\%     & 79.34\%   & 83.23\%  & \textbf{84.91\%}    \\
			Overall Accuracy & 76.50\%     & 79.08\%   & 81.59\%  & \textbf{83.23\%}    \\ 
			\bottomrule
		\end{tabular}
		\label{tabel-6}
	}
\end{table*}

\subsection{Comparisons on the HR Dataset}
Since there are two test sets provided by \cite{lobry2020rsvqa}, we report results on both of them. The experimental results on the HR dataset are displayed in Table \ref{tabel-2} and Table \ref{tabel-3}. 

Table \ref{tabel-2} shows numerical results on the test set 1 of the HR dataset. There are four types of questions in the HR dataset: presence, comparison, count and area. Their corresponding prior weights $W^q_i$ are set as $\{\text{presence: 1.0},$ $ \text{comparison: 3.0}, \text{count: 4.0}, \text{area: 4.0}\}$. The results in Table \ref{tabel-2} reveal that SPCL can consistently enhance the performance of RSVQA for all question types. Particularly, the comparison between baseline method and SPCL indicates that the VQA performance can be improved by simply replacing the cross entropy loss with our SPCL loss. This demonstrates the effectiveness of the proposed training strategy. 

The performance comparison between SPCL and SPCL+MLL shows that the multi-level visual feature is useful for this task. Comparing the baseline method with SPCL+MLL, we also find that improvements for easier question types i.e., comparison and presence, are larger than harder ones. The experimental results reported in Table \ref{tabel-3} demonstrate that the proposed method can outperform the baseline method on all question types on the test set 2 of the HR dataset.

\subsection{Comparisons on the RSIVQA Dataset}
For the RSIVQA dataset, we take MAIN proposed in \cite{zheng2021mutual} as the baseline method. MAIN consists of two modules: a representation module and a fusion module. Image features and question representations are first learned by the representation module. Then, mutual attention and bilinear fusion are utilized to fuse image and question representations in an adaptive manner. The comparison between MAIN and our approach is presented in Table \ref{tabel-4}. On this dataset, according to difficulty levels of question types, prior weights $W^q_i$ are set as $\{\text{yes/no: 1.0},$ $ \text{others: 2.0}, \text{number: 3.0}\}$. Compared with MAIN, our proposed method can achieve much better performance on the yes/no and others types. Although the accuracy of our model on the number type is lower than that of MAIN, the proposed method can obtain better performance in general on the RSIVQA dataset. This demonstrates that the proposed training strategy and multi-level feature learning modules are effective for the RSVQA task.
\subsection{Ablation Study and Discussion}

To evaluate the proposed framework more comprehensively, the following two ablation studies are conducted to explore the effect of different sub-modules. Specifically, to show the superiority of multi-level visual features, CGA and CST modules are compared separately. Quantitative results are presented in Table \ref{tabel-5}. The performance of the model with CGA on three question types is better than that of the baseline method except rural/urban type, which indicates that cross-modal global attention mechanism can be used to enhance the distinguishability of visual features. Since the cross-modal spatial transformer enables more flexible visual feature extraction, the model with CST can achieve better results (except rural/urban type) than both baseline method and the model with CGA. Finally, by combining both modules, MLL model obtains better or competitive performance among these competing methods. This demonstrates the superiority of using multi-level visual features in the RSVQA task.


To further study the effect of different numbers of training samples on model performance, we have conducted experiments by training the proposed SPCL+MLL model with different proportions of the training data. Specifically, 10\%, 40\%, 70\%, and 100\% training samples are used for model training. The results in Table \ref{tabel-6} show that the performance of the model becomes better as the number of training samples increases gradually. In addition, we can also see that the proposed method works well with different sizes of training data.

\section{Conclusion}
\label{Conclusion}
In this paper, two challenges for the RSVQA task are addressed. First, there are no object annotations available in RSVQA datasets, while these annotations can provide rich semantic information for answering questions. Aiming at this challenge, a multi-level visual feature learning method is proposed to jointly learn the language-guided holistic feature and region feature. Specifically, CGA and CST modules are devised to extract flexible visual features with different spatial contexts.
Second, questions in RSVQA datasets are with clearly different difficulty levels for the same remote sensing image. Directly training a model with questions in a random order may confuse the model and limit the performance. In order to alleviate this problem, we propose a language-guided SPCL method with a soft weighting strategy to train networks with samples in an easy-to-hard way. Extensive experiments are conducted on three public RSVQA datasets, and experimental results show that the proposed method can achieve state-of-the-art performance.


\bibliographystyle{IEEEbib}
\bibliography{egbib}

\end{document}